\documentclass[nonacm]{acmart} 

\AtBeginDocument{%
  \providecommand\BibTeX{{%
    \normalfont B\kern-0.5em{\scshape i\kern-0.25em b}\kern-0.8em\TeX}}}

\newcommand{\paragraphbe}[1]{\vspace{.75ex}\noindent{\textbf{#1}} }
\newcommand{\sysname}[0]{\textsc{Clinical Evidence Engine}}
\newcommand{\scenario}[1]{
\noindent\fbox{%
    \centering
    \parbox{\textwidth}{%
        \textbf{Scenario:}~{#1}
    }%
}
\vspace{0.2cm}
\noindent
}

\newcommand{\picoblank}[0]{\rule{1cm}{0.15mm}}

\usepackage{multirow}
\usepackage{array}
\usepackage{siunitx}

\usepackage{graphicx}
\usepackage{wrapfig}
\usepackage{amsmath}
\usepackage{amsthm}
\usepackage{booktabs}
\usepackage{algorithm}
\usepackage{algorithmic}
\usepackage{amsmath,bm,paralist}
\usepackage{comment}

\setcopyright{acmcopyright}
\copyrightyear{2018}
\acmYear{2018}
\acmDOI{10.1145/1122445.1122456}

\acmConference[Woodstock '18]{Woodstock '18: ACM Symposium on Neural
  Gaze Detection}{June 03--05, 2018}{Woodstock, NY}
\acmBooktitle{Woodstock '18: ACM Symposium on Neural Gaze Detection,
  June 03--05, 2018, Woodstock, NY}
\acmPrice{15.00}
\acmISBN{978-1-4503-XXXX-X/18/06}

\begin{document}

\title[\sysname]{\sysname:~Proof-of-Concept For \\ A Clinical-Domain-Agnostic Decision Support Infrastructure}

\author{Bojian Hou}
\email{boh4001@med.cornell.edu }
\author{Hao Zhang}
\email{haz4007@med.cornell.edu}
\affiliation{%
  \institution{Weill Cornell Medicine}
  \city{New York City}
  \state{New York}
  \country{USA}
}

\author{Gur Ladizhinsky}
\email{gur.lad@campus.technion.ac.il}
\author{Ali Kayyal}
\email{ali.kayyal@campus.technion.ac.il}
\affiliation{%
  \institution{Technion}
  \city{Haifa}
  \country{Israel}
}

\author{Stephen Yang}
\email{sy364@cornell.edu}
\affiliation{%
  \institution{Cornell University}
  \city{Ithaca}
  \state{New York}
  \country{USA}
}

\author{Volodymyr Kuleshov}
\email{vk379@cornell.edu}
\affiliation{%
  \institution{Cornell Tech}
  \city{New York City}
  \state{New York}
  \country{USA}
}

\author{Fei Wang}
\email{few2001@med.cornell.edu}
\affiliation{%
  \institution{Weill Cornell Medicine}
  \city{New York City}
  \state{New York}
  \country{USA}
}

\author{Qian Yang}
\authornote{Contact author: Email - \url{qy242@cornell.edu}}
\email{qy242@cornell.edu}
\affiliation{%
  \institution{Cornell University}
  \city{Ithaca}
  \state{New York}
  \country{USA}
}

\begin{abstract}

Abstruse learning algorithms and complex datasets increasingly characterize modern clinical decision support (CDS) systems.
As a result, clinicians cannot easily or rapidly scrutinize the CDSS recommendation when facing a difficult diagnosis or treatment decision in practice.
Over-trusting or under-trusting CDS recommendations are frequent, leading to preventable diagnostic or treatment errors.
Prior research has explored supporting such assessments by explaining DST data inputs and algorithmic mechanisms.
This paper explores a different approach: Providing precisely relevant, scientific evidence from biomedical literature. 
We present a proof-of-concept system, \sysname, to demonstrate the technical and design feasibility of this approach across three domains (cardiovascular diseases, autism, cancer).
Leveraging \textit{Clinical BioBERT}, the system can effectively identify clinical trial reports based on lengthy clinical questions (e.g., "risks of catheter infection among adult patients in intensive care unit who require arterial catheters, if treated with povidone iodine-alcohol").
This capability enables the system to identify clinical trials relevant to diagnostic/treatment hypotheses -- a clinician's or a CDS's.
Further, \sysname~ can identify key parts of a clinical trial abstract, including patient population (e.g., adult patients in intensive care unit who require arterial catheters), intervention (povidone iodine-alcohol), and outcome (risks of catheter infection).
This capability opens up the possibility of enabling clinicians to 1) rapidly determine the match between a clinical trial and a clinical question, and 2) understand the result and contexts of the trial without extensive reading.
We demonstrate this potential by illustrating two example use scenarios of the system.
We discuss future work that can advance the design and ML performances of this system.
We discuss the idea of designing DST explanations not as specific to a DST or an algorithm, but as a domain-agnostic decision support infrastructure.

\end{abstract}



\maketitle

\section{Introduction} 

Biomedical literature can provide valuable ``\textit{decision support}'' for clinicians. From best practices to clinical trial reports, the literature contains scientifically proven information that can aid numerous diagnostic and therapeutic dilemmas across all clinical domains.
With the rise of Evidence-Based Medicine (EBM), clinicians increasingly turn to literature at point-of-care to inform their decisions~
\cite{rosner2012evidence, smith1996clinical};
Medical students receive training on how to formulate their clinical questions into good search terms, for example, training on the PICO (Population, Intervention, Comparison, and Outcome) clinical knowledge representation framework~\cite{masoomi2012best}.

Interestingly, literature is rarely in the spotlight of clinical decision support (CDS) research. 
With the explosive growth in machine learning (ML), CDS systems are increasingly characterized by patient-data-driven inferences and abstruse risk models, each tailored for one clinical decision or one clinical domain. 
In this context, literature-based systems can appear particularly valuable for clinicians today. They can complement other CDS systems and offer scientific evidence that clinicians can easily understand~\cite{elwyn2013many, Yang_unremarkable_AI,cai2019hello}. They can support many vastly different clinical decisions and domains, including those in data-poor or resource-constrained hospitals.

Document-level literature retrieval systems such as PubMed and Google Scholar have proven their worth in clinical practice. 
However, at points-of-care, clinicians need far more fine-grained tools to identify information that is \textit{precisely applicable} to their patient case and clinical question at hand~\cite{ebm-manual,masoomi2012best}.
Clinical decision-making is a continuous and iterative process, consisting of a series of micro-decisions \cite{dst-framework-tiffen2014,bate2012clinical,yangVAD2016}.
For each micro decision in practice, clinicians could only afford 2-3 minutes on literature search~\cite{del2014clinical}.  
To be useful, literature systems need to be able to directly answer point-of-care clinical questions such as``\textit{What are the risks of catheter infection if an adult patient in intensive care unit who requires arterial catheters is treated with povidone iodine-alcohol rather than chlorhexidine–alcohol}?''~\cite{masoomi2012best}
Existing systems (e.g., PubMed, \textit{Trialstreamer} \cite{trialstreamer-mapping}) struggle with processing such complex queries.
Searching the query above on PubMed simply returns no results, not to mention eliciting key information from the literature.

This work aims to more precisely address point-of-care clinical questions with biomedical literature.
We present \sysname, a proof-of-concept system that demonstrates the technical feasibility of achieving this goal.
On clinical trial reports of three domains (cardiovascular diseases, autism, cancer), \sysname~can: \textbf{(Capability 1)} Identify relevant clinical trial reports based on complex, long, clinical-question-like search queries (up to 512 words), queries such as ``\textit{Would the addition of radiotherapy on top of androgen-deprivation therapy lead to higher risk of bowel toxicity of an adult male patient with locally advanced prostate cancer?}''
The literature retrieval model of \sysname~achieves an accuracy of 99.44\%, when evaluated on synthetic queries based on an established expert-annotated literature dataset \cite{nye_trialstreamer_2020}.
\textbf{(Capability 2)} Identify critical information in clinical trial report abstracts can serve as clinical evidence, i.e., PICO (Population, Intervention, Comparison, and Outcome) information \cite{masoomi2012best}. The PICO classification model of \sysname~achieves a F1 score of 0.74. Both models outperform existing state-of-the-art models.

These newfound technical capabilities open up new design and research opportunities around literature-based CDS; around designing clinical decision supports as domain-agnostic, intelligent information infrastructure, rather than decision- or domain-specific applications.
We concretize these opportunities and questions through two example use scenarios of \sysname~: aid clinicians in crutinizing (i) their self-derived decision hypothesis and (ii) an abstruse patient-data-based risk model.

This paper makes two contributions.
First, it demonstrates novel bioNLP capabilities of harnessing biomedical literature as point-of-care decision.
Key to this technical advance is the integrative use of a clinical knowledge representation framework (i.e. PICO~\cite{masoomi2012best}) and large pretrained language models (i.e. Clinical BioBERT \cite{DBLP:journals/corr/abs-1904-03323}).
Second, \sysname~offers an initial design exemplar of a domain-agnostic, intelligent information infrastructure. It offers an alternative perspective to the traditional idea of clinical decision supports as decision- or domain-specific applications. It can serve as a valuable point of reference for the research discourse on future AI-infused healthcare.

\section{Related Work} 
\subsection{Biomedical Literature Supports Clinical Decision-Making}

Clinicians routinely consult biomedical literature for decision-making. When facing diagnostic and prognostic dilemmas, clinicians search the literature -- most often on PubMed -- for valuable clinical evidence such as clinical trial reports, best practices, expert-annotated case studies, and biological explanations \cite{ebm-manual,lu2011pubmed,kendall2017pubmed}.
Such evidence compliments clinician judgments and patient value, forming the cornerstones of Evidence-Based Medicine \cite{EBM-workbook}.

At points of care, clinicians search literature for \textit{highly-specific} information to address the clinical question at hand. 
Such searches are often very challenging given the overwhelming amount of literature available \cite{ebm-manual}.
To address this challenge, clinicians and medical students receive mandatory trainings on how to translate messy clinical situations into effective literature search queries, for example, by using the PICO framework~\cite{masoomi2012best,PICO-table}.

\begin{table}[h]
    \centering
    \begin{tabular}{|p{3cm}|p{11cm}|}
        \hline
        \textbf{Question Type} & \textbf{PICO Template For Formulating Clinical Questions}\\
        \hline
        Therapy/intervention & In \picoblank (\textbf{P}opulation), what is the effect of \picoblank (\textbf{I}ntervention), compared with \picoblank (\textbf{C}omparator) on \picoblank (\textbf{O}utcome)? \\
        Diagnosis/assessment & For \picoblank (population), does \picoblank (tool/procedure) yield more accurate or appropriate diagnostic/assessment information than \picoblank (comparator tool/procedure) about \picoblank (outcome)? \\
        Prognosis & For \picoblank (population), does \picoblank~(disease/condition) relative to \picoblank (comparator disease/condition) increase the risk for \picoblank (outcome)? \\
        \hline
    \end{tabular}
    \vspace{0.2cm}
    \caption{Part of the PICO (Population, Intervention, Comparison, and Outcome), a clinical knowledge representation framework \cite{masoomi2012best,PICO-table}. Clinicians receive training on how to use such frameworks to effectively search biomedical literature for clinical evidence.}
    \label{tab:PICO}
\end{table}

Besides specificity, standards of \textit{rigor} also drive clinicians' literature search.
Healthcare communities differentiate the levels of evidence in biomedical literature~\cite{levels-evidence-pyramid}. They consider population-level evidence (e.g., randomized controlled trials and systematic reviews) more rigorous and trustworthy than isolated observations (e.g., peer-reviewed case studies)~\cite{rosner2012evidence,smith1996clinical}.
The importance of rigor is also evident in clinicians' choice of literature search tools.
In research, clinicians built expert-curated-and-maintained literature databases~\cite{fox2003uptodate,journalclub}.
In practice, more clinicians use PubMed (a keyword-matching-based search engine) than Google Scholar. Despite higher accuracy, the latter is considered less rigorous, as its rankings consider article popularity~\cite{shultz2007comparing, shariff2013retrieving}. 

Empirical research is sparse on how clinicians use literature in practice.
This is in sharp contrast with nonmedical domains, where sense-making and HCI research have extensively studied people's organic information search/foraging and decision-making. They created tools to support sense-making, such as novel visualization techniques and crowd-sourced knowledge representation structure \cite{chang_searchlens_2019, chang_mesh_2020,pirolli_sensemaking_nodate}.

\subsection{Clinical Decision Support and Explainable AI in HCI Research}

Computational decision support (CDS) systems are systems that assist point-of-care decision-making. 
They promise to improve patient outcomes~\cite{moghadam2021effects}, reduce medical error~\cite{jia2014literature}, and reduce healthcare disparities~\cite{amirfar2011developing}.
In recent years, with the increasing digitalization of Electronic Health Records (EHR) and the explosive growth in machine learning (ML), clinician-facing CDS research is increasingly characterized by complex algorithms and patient-data-driven inferences~\cite{kuperman2007medication,nenova2021chronic}. 
Since the focus of this work is on biomedical literature, we only briefly review this body of work, particularly the challenges it has reported. This is not meant to be a criticism. Instead, these challenges motivated us to use biomedical literature to \textit{complement} non-literature-based CDS systems. More comprehensive reviews of CDS work and their achievements can be found elsewhere \cite{elwyn2013many,wyatt1995commentary,yang2015review,nenova2021chronic}.

\begin{itemize}
    \item \textit{Interpretability and accessibility to clinicians.}  Abstruse, data-driven algorithms increasingly characterize modern CDS. Clinicians often found these systems too time-consuming to understand, their explanations overly complicated \cite{devaraj2014barriers,Yang_unremarkable_AI,why-expert-systems-fail-1985}. Over-trusting or under-trusting CDS recommendations are frequent, leading to preventable diagnostic or treatment errors \cite{xie2020chexplain,Yang_unremarkable_AI,overgrade2012,GP5undergrade2011}. 
    
    \item \textit{Sense of rigor.}~Clinicians did not always trust the diagnostic/treatment predictions, even when they fully understand how the ML correctly generated the prediction~\cite{xie2020chexplain,Yang_unremarkable_AI}. This is because the standards of rigor for clinical evidence are often at odds with the basis on which the rigor of ML is premised. For example, inference-based diagnostic predictions do not qualify as ``\textit{empirical evidence}'' under the levels of evidence pyramid \cite{levels-evidence-pyramid}. Empirical research reported that, when judging the trustworthiness of CDS, clinicians asked whether it has been published in top-tier clinical journals~\cite{Yang_unremarkable_AI}. 
    
    \item \textit{Generalizability challenges.} 
    Researchers most often tailored data-driven CDS models and designed explanations for particular clinical decisions, data types, and/or algorithms~\cite{xie2020chexplain, cai2019hello}.
    Given the multitude of clinical decisions involved in caring for each patient, it can seem that, at point-of-care, future clinicians will need to make sense of multiple CDS predictions in quick succession (e.g., blood-test-based diagnostic support, computer vision-based X-ray reading, tabular-disease-trajectory prediction, etc.).
They are also responsible for scrutinizing each prediction and accounting for its potential biases. This will be extremely difficult.
\end{itemize}

Within such a milieu, the inherent characteristics of literature -- \textit{scientifically-proven, domain-agnostic, accessible} -- can be particularly valuable for clinicians today \cite{elwyn2013many, Yang_unremarkable_AI,cai2019hello}. 

\subsection{Mining Biomedical Literature in BioNLP Research}\label{bioNLP-relatedwork}

Algorithmically retrieving and processing biomedical literature are challenging tasks at the frontier of NLP research. In comparison to other documents, biomedical literature includes complex terminologies, concepts, and relationships that even scientists may not fully understand~\cite{neveol2018clinical}. Popular literature mining systems, such as PubMed, utilize keyword matching techniques. They can struggle with many search queries that clinicians need at point-of-care~\cite{kendall2017pubmed}.

Large pretrained language models promise substantial advances on these fronts. For example, BERT~\cite{DBLP:journals/corr/LillicrapHPHETS15} can perform literature mining tasks such as documents retrieval and key information extraction. \textit{Clinical BioBERT}~\cite{lee2020biobert} can even more effectively capture biomedical and clinical knowledge as it has been fine-tuned with biomedical and clinical documents.
Literature datasets created for enhancing Evidence-Based Medicine (EBM) can also catalyze novel literature mining and information retrieval capabilities. One such dataset is the EBM-NLP dataset~\cite{DBLP:conf/acl/NenkovaLYMWNP18}, a corpus of $4991$ clinical trial report abstracts annotated with PICO elements. The abstracts are extracted from PubMed and focus on cardiovascular diseases, cancer, and autism. Medical experts and crowd-workers collaboratively annotated the PICO elements. These PICO annotations can serve as the ground truth for many EBM-related NLP research efforts. 

Researchers have started to leverage these large language models and datasets in creating novel biomedical literature mining systems.
For example, a COVID-19 literature mining system that can surface emergent research directions on the topic~\cite{covid-scientific-directions-2021} and a question answering systems also on scholarly COVID literature~\cite{su2020cairecovid}.
The most closely related to point-of-care is TrailStreamer~\cite{marshall2020trialstreamer}, which uses both ML and rule-based methods to extract PICO information from human-subject Randomized controlled trials (RCTs) reports. While advancing on classification performance, Trialstreamer has limited success in retrieving literature with long or highly-specific queries. For example, Trialstreamer can only define a patient population based on a single clinical condition (e.g., all patients with diabetes). This limits the system from identifying RCTs that match more meaningfully with patient cases, e.g., according to their medical histories, commodities, and demographics.

\section{System Design Goals and Overview}

We set out to create a literature-based system that can support clinicians' point-of-care decision-making. Drawing up prior work, we had two goals: (1) To process long, complex clinical questions as literature search queries, e.g., ``\textit{what are the risks of catheter infection if an adult patient in intensive care unit who requires arterial catheters is treated with povidone iodine-alcohol rather than chlorhexidine–alcohol?}''; (2) To identify not only relevant literature documents but the clinically-relevant information within.
These orientations require state-of-the-art bioNLP capabilities.
They also differ from the convention
of CDS designs, in which each system is tailored for particular clinical decisions or domains.

\begin{figure}[h]
  \centering
  \includegraphics[width=0.9\linewidth]{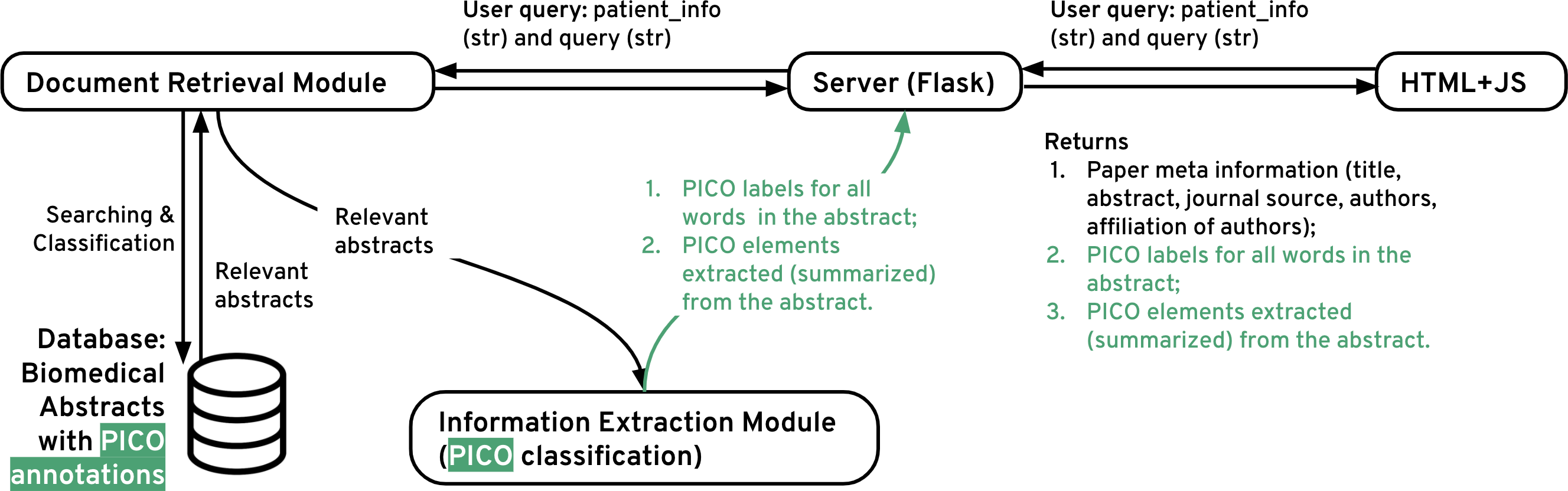}
  \caption{\sysname~system architecture. Key to its design is the organic integrative use of large pretrained language models (i.e. Clinical BioBERT~\cite{DBLP:journals/corr/abs-1904-03323}) and a clinical knowledge representation framework (i.e. PICO \cite{masoomi2012best}, highlighted in green.}
  \Description{System architecture design.}
  \label{fig:system architecture}
\end{figure}

Towards these goals, we designed a system architecture that integrated both large pre-trained language models (i.e. Clinical BioBERT~\cite{DBLP:journals/corr/abs-1904-03323}) and a clinical knowledge representation framework (i.e. the PICO framework and annotations \cite{masoomi2012best,DBLP:journals/corr/abs-1904-03323}.)
The former offers the capabilities to process long, complex biomedical texts across domains (up to 512 words); The latter allows us to identify clinically relevant information from the literature. Figure~\ref{fig:system architecture}
illustrates the system architecture design.

\sysname~'s backend includes two modules. One is the \textit{Document Retrieval Module}, which retrieves relevant biomedical literature articles according to long, complex clinical questions as search queries. To the best of our knowledge, no prior work has created such retrieval models. 
The other is the \textit{Information Extraction Module}, which identifies and extracts the PICO elements within the article's abstract. Prior research reported challenges in balancing such models' precision and recall, thus causing relatively low F1 scores~\cite{DBLP:conf/acl/NenkovaLYMWNP18}. We aim to address this challenge.

Given our focus on biomedical literature, we chose to train our models using the EBM-NLP dataset~\cite{DBLP:conf/acl/NenkovaLYMWNP18}, a clinical trial report dataset with expert annotations of PICO elements for each report. It focuses on three clinical domains: cardiovascular disease, autism, and cancer.  As a result, our system will also focus on trial reports on these domains.

\section{Developing a Proof-of-Concept System}\label{sectionML}

\subsection{Point-of-Care Biomedical Literature Retrieval}

\begin{figure}[h]
  \includegraphics[width=0.9\linewidth]{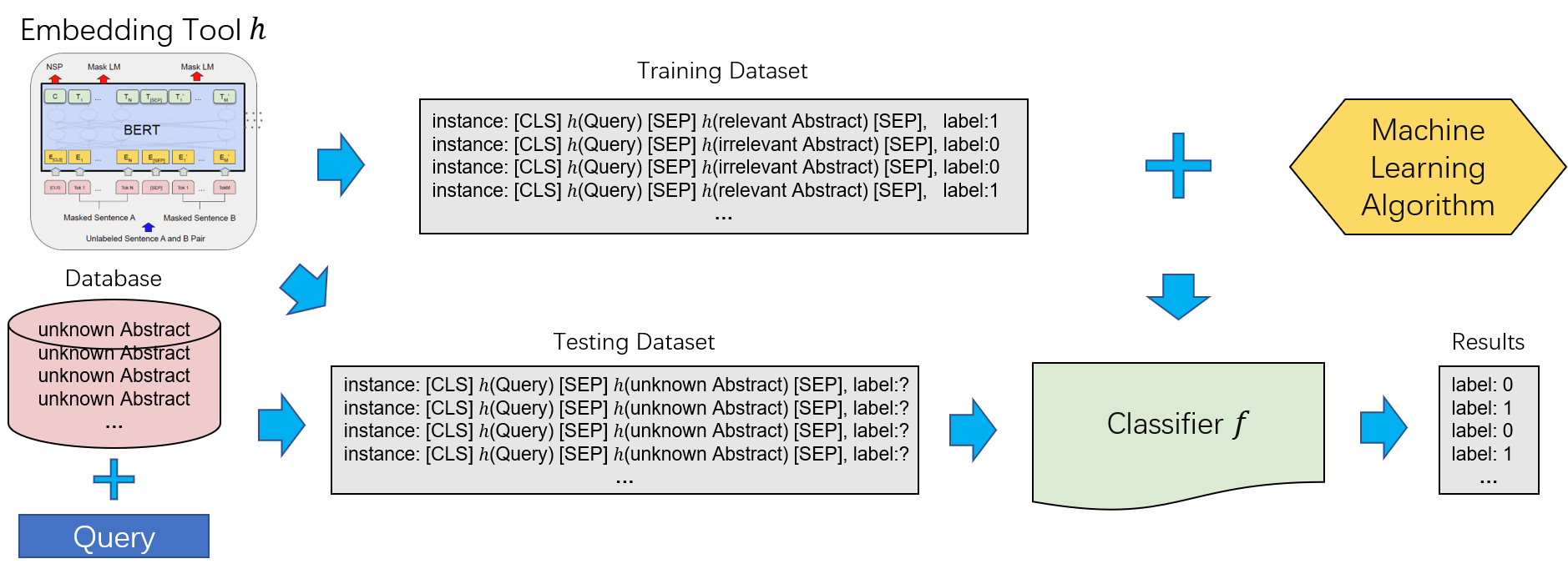}
  \caption{Workflow of the document retrieval module.}
  \label{fig:retrieval model}
\end{figure}

We set out to train a document retrieval model by fine-tuning the Clinical BioBERT models on concatenated \emph{(query, abstract)} pairs: 
\begin{equation}
\text{[CLS]}\ h\text{(Query) [SEP]}\ h\text{(Abstract) [SEP]}
\end{equation}
where $h$ is the BERT model, ``[CLS]'' and ``[SEP]'' are the special signs for the input of BERT.
At test time, we used this model to predict the probability that a query is associated with each abstract in the dataset. The model then returns a literature document ranking based on the probability scores.
Fig.~\ref{fig:retrieval model} illustrates this process in detail.

One critical challenge, however, is that the EBM-NLP dataset does not include search queries for such training.
To address this challenge, we generated synthetic search queries by concatenating the expert-annotated PICO elements in literature abstracts.
As a result, each literature abstract represents a perfect match with one synthetic search query. The match represents one positive training instance ``[CLS] $h$(Query) [SEP] $h$(\textbf{relevant} Abstract) [SEP]'', while others represent negative instances ``[CLS] $h$(Query) [SEP] $h$(\textbf{irrelevant} Abstract) [SEP].''
These synthetic queries can effectively simulate clinician search queries, because the PICO framework is the best practice with which clinicians formulate point-of-care clinical questions into search questions \cite{masoomi2012best,PICO-table}.

\begin{figure} [t]
  \includegraphics[width=0.6\linewidth]{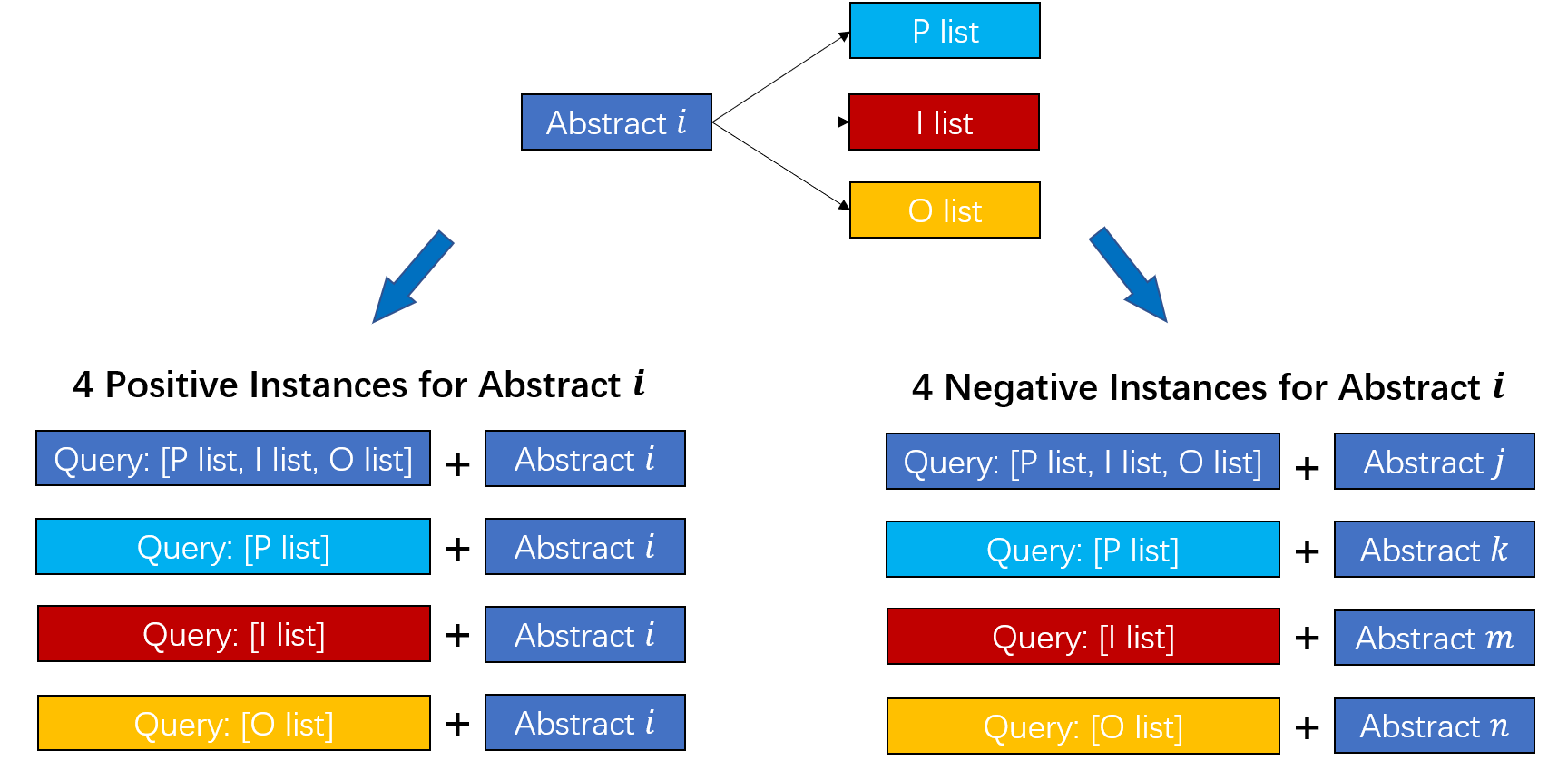}
  \caption{The generation process of the retrieval model dataset, where $i\neq j\neq k\neq m\neq n$. For abstract $i$, we first extract its PICO elements and then generate the positive and negative instances with queries of different lengths.}
  \label{fig:dataset generation for retrieval}
  \vspace{0.2cm}
\end{figure}

Clinical questions in practice do not always include all PICO elements, for example, some diagnostic questions do not have a comparator \cite{PICO-table}. In this light, we generated synthetic queries that included all PICO elements as well as those that only included a subset. This data generation process (Fig.~\ref{fig:dataset generation for retrieval}) generated 4 positive instances and 4 negative instances for each abstract.

\begin{table}[t]
    \begin{tabular}{|p{2cm}|p{2cm}|p{4.5cm}|p{4.5cm}|}
		\hline
		 & \textbf{Accuracy} & \textbf{F1 for Negative Relevance} & \textbf{F1 for Positive Relevance} \\ 
		 \hline
		 This work & 0.9944$\pm$.00089 & 0.9945$\pm$.00091 & 0.9944$\pm$.00087 \\
		 Best Baseline & 0.9535 & 0.9549 & 0.9519 \\
		 \hline
	\end{tabular}
	\vspace{0.2cm}

	\begin{tabular}{|p{2cm}|p{2cm}|p{4.5cm}|p{4.5cm}|}
		\cline{3-4}
		\multicolumn{2}{c|}{} & \textbf{Predicted Negative Relevance} & \textbf{Predicted Positive Relevance}\\ 
		 \hline
		\multirow{2}{*}{This work} &True Negative &  3940 & 24 \\ 
		&True Positive & 21 & 3943  \\ 
		 \hline
		\multirow{2}{*}{Best Baseline} &True Negative &  3905 & 59 \\ 
		&True Positive & 310 & 3654  \\ 
		 \hline
	\end{tabular}
	\vspace{0.2cm}
	\captionsetup{width=0.7\textwidth}
	\caption{Top: Retrieval model accuracy and F1 with standard deviation over five runs. Bottom: Retrieval model confusion matrix (one run).} 
	\label{tab:LReval}
\end{table}
Next, we randomly selected $4000$ abstracts for training and used the rest $991$ for testing. We trained the retrieval model on the $4000\times8=32000$ training instances and evaluate it on the $991\times8=7928$ testing instances. We run $5$ times with different splittings of the training and testing datasets.

This evaluation process revealed a F1 score of 0.9945 for positive document relevance, and 0.9944 for negative document relevance.
It also shows that quantitatively, our model outperforms the best results of the keyword matching approach \footnote{In our experiment, keyword-based approaches achieved their best performance when the system retrieves only the documents that include 40\% of the keywords appeared in a search query. This performance is inferior to our models, as detailed in Table~\ref{tab:LReval}.}, a common strategy used by popular literature retrieval systems.
Table~\ref{tab:LReval} details the performance comparison between this work and the best results from keyword matching approaches, in terms of accuracy, F1 score, confusion matrix.
One limitation of this evaluation is that it focused solely on whether the system possesses the ability to identify all relevant documents. Confusion matrix, accuracy and F1 score are sufficient to measure this ability. However, it did not assess the model's ranking abilities (e.g., calculating the NDCG or AUC values). Unfortunately, there exists no ground truth datasets that could enable such evaluation. 

\subsection{Salient Information (PICO) Extraction Module}

\begin{figure}[!t]
  \includegraphics[width=0.9\linewidth]{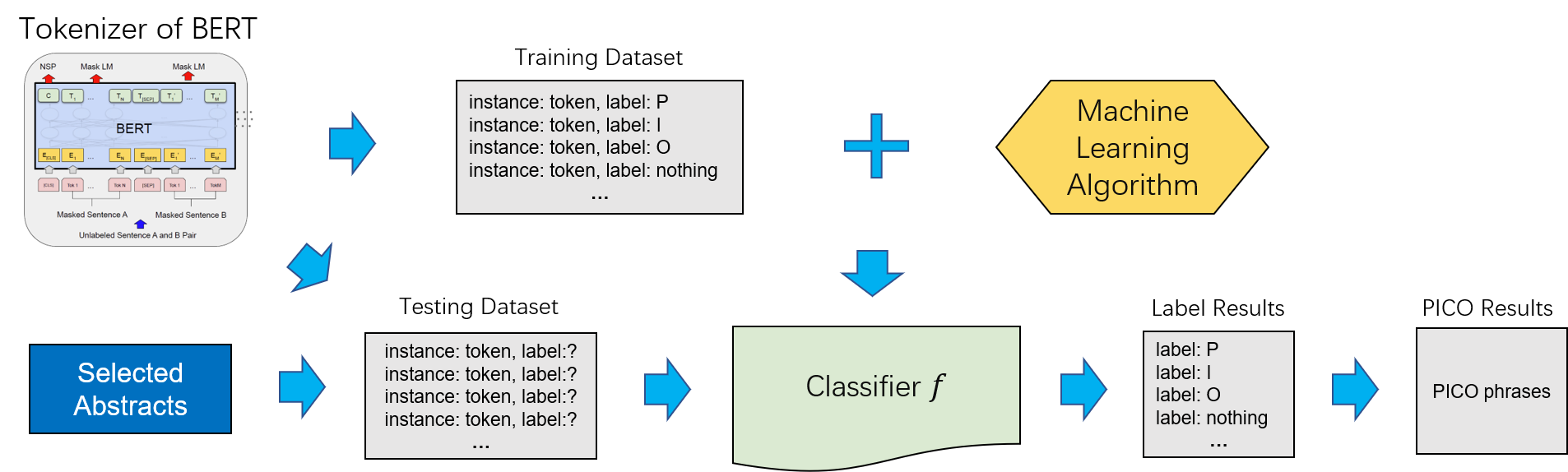}
  \caption{Illustration of information extraction module.}
  \vspace{0.2cm}
  \label{fig:PICO classification model}
\end{figure}

The information extraction module aims to identify and summarize the salient elements of clinical evidence from literature abstracts. Based on the PICO framework, we consider Population (P), Intervention/Comparator (I/C), and Outcome (O) as salient information \cite{PICO-table,masoomi2012best}.
We used Clinical BioBERT tokenizer to tokenize the words into tokens that are expressed as numerical vectors. Using ``(token, label)'' pairs as training data, we trained a linear four-class classifier with the EBM-NLP dataset; a classifier that predicts whether each token in the abstract describes P, I/C, or O.
After obtaining the classification results, the system groups adjacent words with same annotations and remove duplicates to generate the final PICO phrases.
Fig.~\ref{fig:PICO classification model} summarizes the information extraction module workflow.

We evaluated this model and compared it to the three state-of-the-art PICO classifiers in~\cite{DBLP:conf/acl/NenkovaLYMWNP18}. These include logistic regression (LogReg), LSTM-CRF~\cite{lample2016neural} and LSTM-CRF-BERT, which uses BERT as the embedding tool for LSTM-CRF.
For fair comparison, we tested all four models on the extra 200 withheld testing data from EBM-NLP~\cite{DBLP:conf/acl/NenkovaLYMWNP18}.
Our model outperforms other methods in F1 score, achieving balanced precision and recall (Figure~\ref{tab:PICO Classification}.) 

\begin{table}[!t]
	\centering
	\setlength{\tabcolsep}{5mm}
	\begin{tabular}{|c|c|c|c|}
		\hline
		 \textbf{Model} & \textbf{F1} & \textbf{Precision} & \textbf{Recall}\\ 
		 \hline
		 LogReg &  $0.45$ & $0.31$ & $0.82$ \\ 
		 LSTM-CRF & $0.68$ & $0.70$ & $0.66$ \\ 
		 LSTM-CRF-BERT & $0.68$ & $0.69$ & $0.66$   \\ 
		 This Work & $0.73$ & $0.70$ & $0.76$\\
		 \hline
	\end{tabular}
	\vspace{0.2cm}
	\captionsetup{width=.75\textwidth}
	\caption{Key information (PICO) extraction model performance. Leveraging Clinical BioBERT, our model outperforms state-of-the-art methods on F1 score.}
	\label{tab:PICO Classification}
\end{table}

\section{Example Interaction Scenarios:\\Design and Research Opportunities in Harnessing Literature as Decision Support}

We have so far described how \sysname~demonstrates two novel technical capabilities necessary for harnessing biomedical literature as point-of-care decision support: (1) Identifying relevant literature documents based on long, complex search queries and (2) extracting precisely relevant information (i.e., PICO elements) from the identified documents.
The newfound technical capabilities mark a clear design space in supporting biomedical literature sense-making. They also open up new HCI research opportunities around supporting clinical decision-making \textit{across domains} with literature.
To concretize these design and research opportunities, below we discuss them via two example use scenarios of \sysname~: helping clinicians to (i) answer their self-derived clinical questions and to (ii) scrutinize a deep-learning-based risk model.

\subsection{Use Scenario 1: Supporting Clinicians  Questions as Search Queries}

\scenario{The clinician team just diagnosed a male adult patient with prostate cancer. At that point, the cancer cells have only been developing locally. Based on their clinical acumen and standard practice, the clinicians have decided to deploy androgen-deprivation therapy (ADT). However, they are unsure, for this particular patient, whether the addition of radiation therapy (RT) would further improve their chance of survival. This is a critical decision, as radiation has substantial side effects and should not be used casually.}

\noindent
Facing this therapeutic dilemma, clinicians start searching for relevant trials on \sysname~using a nuanced description of the scenario as search term.
\begin{quote}
    \textbf{Clinician's search query:}~\textit{Would the addition of ``radiotherapy'' on top of ``androgen-deprivation therapy'' help improve the survival of an adult male patient with ``locally advanced prostate cancer''?}
\end{quote}

\begin{figure}[t]
  \includegraphics[width=\linewidth]{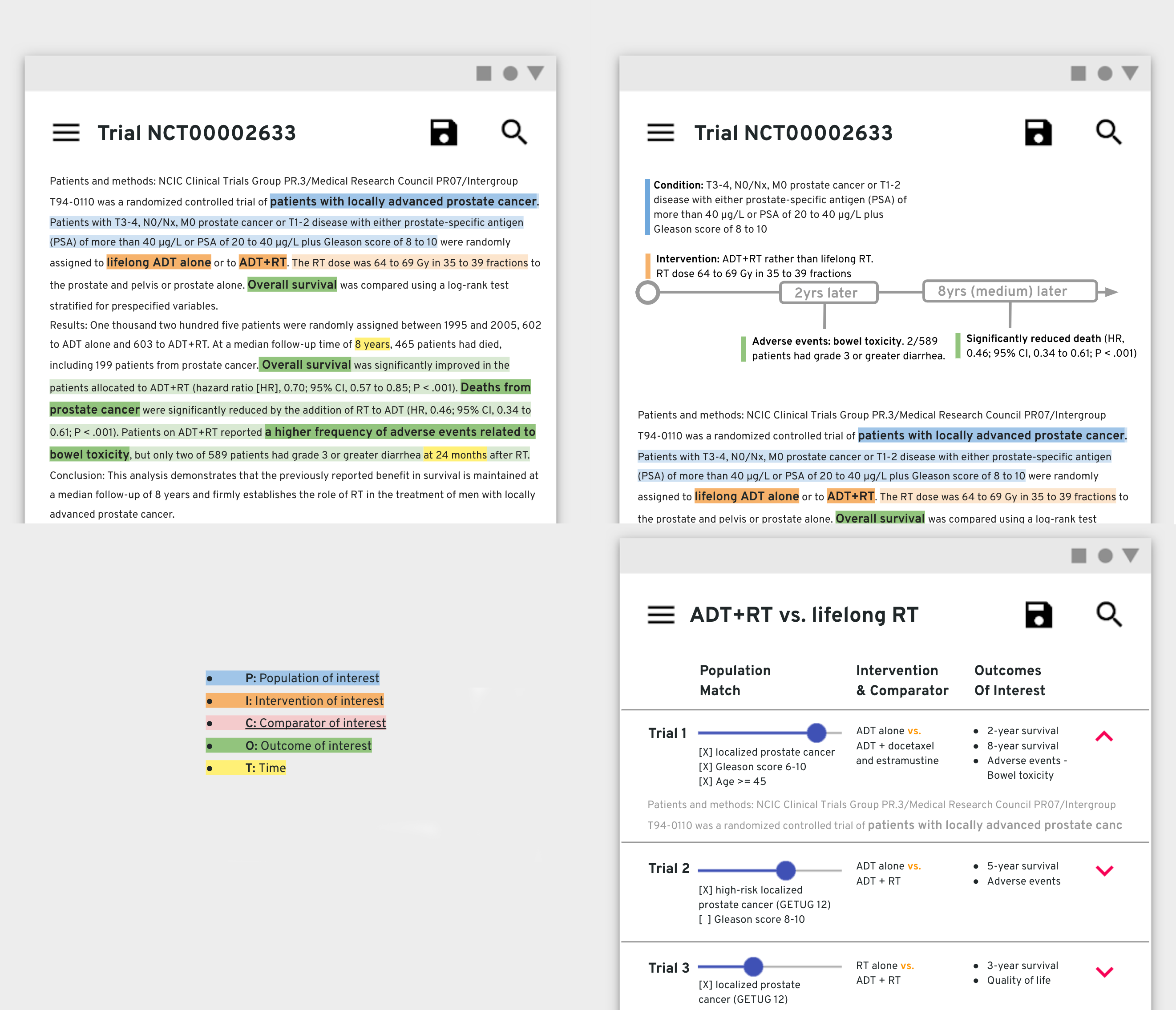}
  \caption{Example interface designs of \sysname~as it directly addresses clinicians' points-of-care clinical questions. Each embodies a sense-making support design lesson in prior HCI research; All are enabled by the ML capabilities described in section \ref{sectionML}.}
  \label{fig:scenario1}
\end{figure}

\paragraphbe{Technical capabilities.~}
Based on this query, \sysname~can identify a list of relevant randomized controlled trial reports. 
The highest ranking result is the report ``\textit{Final report of the intergroup randomized study of combined androgen-deprivation therapy plus radiotherapy versus androgen-deprivation therapy alone in locally advanced prostate cancer}''~\cite{mason2015final}. It precisely matches the clinical question in terms of clinicians' population, interventions, and outcome of interest. Further, the PICO classifier of \sysname~extracts the following information and can concisely answer the clinical question raised.

\begin{itemize}
    \item \textbf{P}opulation: ``\textit{patients with locally advanced prostate cancer}'', and more specifically, ``\textit{Patients with T3-4, N0/Nx, M0 prostate cancer or T1-2 disease with either prostate-specific antigen (PSA) of more than 40 $\mu$g/L or PSA of 20 to 40 $\mu$g/L plus Gleason score of 8 to 10}'';
    
    \item \textbf{I}ntervention and \textbf{C}omparator:
    ``\textit{Combined Androgen-Deprivation Therapy Plus Radiotherapy Versus Androgen-Deprivation Therapy Alone}''. specifically, ``\textit{lifelong ADT alone}'';
    
    \item \textbf{O}utcome: ``\textit{overall survival}'', ``\textit{deaths from prostate cancer}'', and ``\textit{frequency of adverse events related to bowel toxicity}''.
\end{itemize}

\paragraphbe{Design opportunities.~}How can literature-based CDS systems best support clinicians' literature sense-making and clinical decision-making with the retrieved PICO elements? Extensive prior HCI work has studied how to support information foraging, sense-making, and decision-making~\cite{chang_mesh_2020,chang_searchlens_2019}.
\sysname~'s novel technical abilities illuminate a clear design space for expanding this research into biomedical domains. 
For example, prior research has shown that scaffolding search process and results can reduce users' cognitive efforts, enabling them to build up a deeper understanding of the decision being made~\cite{chang_mesh_2020}. PICO information extraction capabilities enable such scaffolding, for example, by allowing clinicians to compare populations and outcomes on comparable interventions across studies (Figure~\ref{fig:scenario1} bottom right).
Prior research has also built tools that allow users to create a collection of composable and reusable ``lenses'' to reflect their different latent interests \cite{chang_searchlens_2019}. Such tools were effective in improving users' depth of information understanding. Future literature-based CDS tools can explore enabling clinicians to create dynamic and reusable lenses. In this particular search scenario, clinicians may find the patient-population-match lens valuable, as they can use it to identify the trials that most closely align with the patient at hand in terms of cancer type, severity, and spread. Later in the search process, clinicians can shift to a temporal lens, rapidly and effectively examining the temporal progression of ADT-plus-radiotherapy treatment effects (Figure top right).
Finally, visualizing the key clinical evidence in the literature exemplifies near-term, pragmatic design solutions. It can offer perceptual cues for clinicians and expedite information foraging at points of care~\cite{information-scent} (Figure top left). 

\paragraphbe{Open research opportunities.~}PICO represents only one way of summarizing salient clinical evidence in biomedical literature. Future research should advance on the machine learning models described in this work by systematically investigating other ontological clinical information frameworks (e.g., levels of evidence) and assessing their effectiveness in point-of-care decision support. Other summarization or knowledge extraction techniques beyond supervised learning can also be valuable. There is a real need for curating HCI- and CDS-oriented biomedical literature datasets. It is critical for such human-centered ML research to advance further. 

Clinicians not only need concise and precisely relevant clinical evidence, but they also need enough contextual integrity to interpret and act on this evidence. So far, little to no empirical research has studied how clinicians forage and utilize clinical evidence from literature \textit{in practice}. 
This should be a critical research question in CDS design and research.
It can help illuminate clinicians' sequential search behaviors and high-level strategies, further improving the retrieval and extraction performances.
It can also inform summarization or visualization design strategies and ensure the correct interpretation of extractive literature evidence for point-of-care decision-making. 


\subsection{Use Scenario 2: Harnessing Literature Along with Other Decision Support Tools}


\scenario{A nurse practitioner in a small, rural hospital is assessing whether a 15-month-old girl with disruptive behavior history has Autism Spectrum Disorder (ASD). She does so by conducting the standard Autism Diagnostic Interview (ADI), Autism Diagnostic Observation Schedule (ADOS), as well as observing her interaction with her family members and strangers in the hospital. While the standard tests (ADI and ADOS) indicate autism, the nurse practitioner notices that the girl has a secure attachment relationship with her parents. Does that mean this girl does not have autism, considering that autistic children most often exhibit pervasive deficits in social, affective, and communicative behaviors? The nurse practitioner wonders.

\textbf{Other decision support systems at play.~}In this context, the nurse practitioner looks up her decision support systems. One autism detection CDS she uses\cite{tariq2018mobile} analyzes a 3-minute video of the girl interacting with her family. It predicts that the girl is 54\% likely to have ASD. This system is trained on large video databases of children with and without ASD and has a 92\% accuracy. However, the system deploys eight complex ML models to make a diagnostic prediction collectively. The nurse practitioner struggles to decide whether to trust this prediction or not.
}

\noindent
Facing this diagnostic dilemma, the nurse practitioner opens up \sysname~and gives it permission to use this patient's medical industry and the featurized video data that the autism detection CDS has produced.
\begin{quote}
    \textbf{Auto-generated search query:}~\textit{Female child, ``disruptive behaviors'', ``secure parental attachment'', ``Autism Spectrum Disorder'', ``social behavior after separation from parents.''}
\end{quote}

\paragraphbe{Technical capabilities.~}
Based on this query, \sysname~identifies a list of relevant biomedical literature documents.
The highest ranking result, a publication named ``\textit{A Parent-Mediated Intervention That Targets Responsive
Parental Behaviors Increases Attachment Behaviors in Children
with ASD: Results from a Randomized Clinical Trial}'', which largely matches the clinical question in terms of clinicians' population, interventions, and outcome of interest. 
More importantly, the system extracted the P,I, and O elements that together read ``\textit{[...]the attachment behaviors of children with Autism Spectrum Disorder (ASD) show striking similarities to those of typically developing children.}'' -- A sentence that addresses the nurse practitioner's question and confirms a positive diagnosis.


\begin{itemize}
    \item \textbf{P}opulation: ``\textit{children aged 12 and 24 months diagnosed with ASD}'', ``children with Autism Spectrum Disorder (ASD) ''.
    \item \textbf{I}ntervention: ``\textit{Focused Playtime Intervention (FPI)}'', a type of ``\textit{parent-mediated intervention}''
    \item \textbf{C}omparator: n/a
    
    \item \textbf{O}utcome: ``\textit{parental perceptions of child attachment}'', ``\textit{attachment related outcomes}'', ``\textit{attachment-related behaviors}'', ``\textit{similarities to those of typically developing children}''
\end{itemize}



\paragraphbe{Design opportunities.~}
Abstruse algorithms and complex patient-data-driven inferences increasingly characterize modern CDS systems. Clinicians have frequently reported challenges in understanding the trustworthiness of such systems and their predictions, especially under the time constraints of busy clinical work \cite{devaraj2014barriers,Yang_unremarkable_AI,why-expert-systems-fail-1985}.
The multi-learning-algorithm, video-based autism detection system offers merely one example.

\begin{figure}[ht]
  \includegraphics[width=\linewidth]{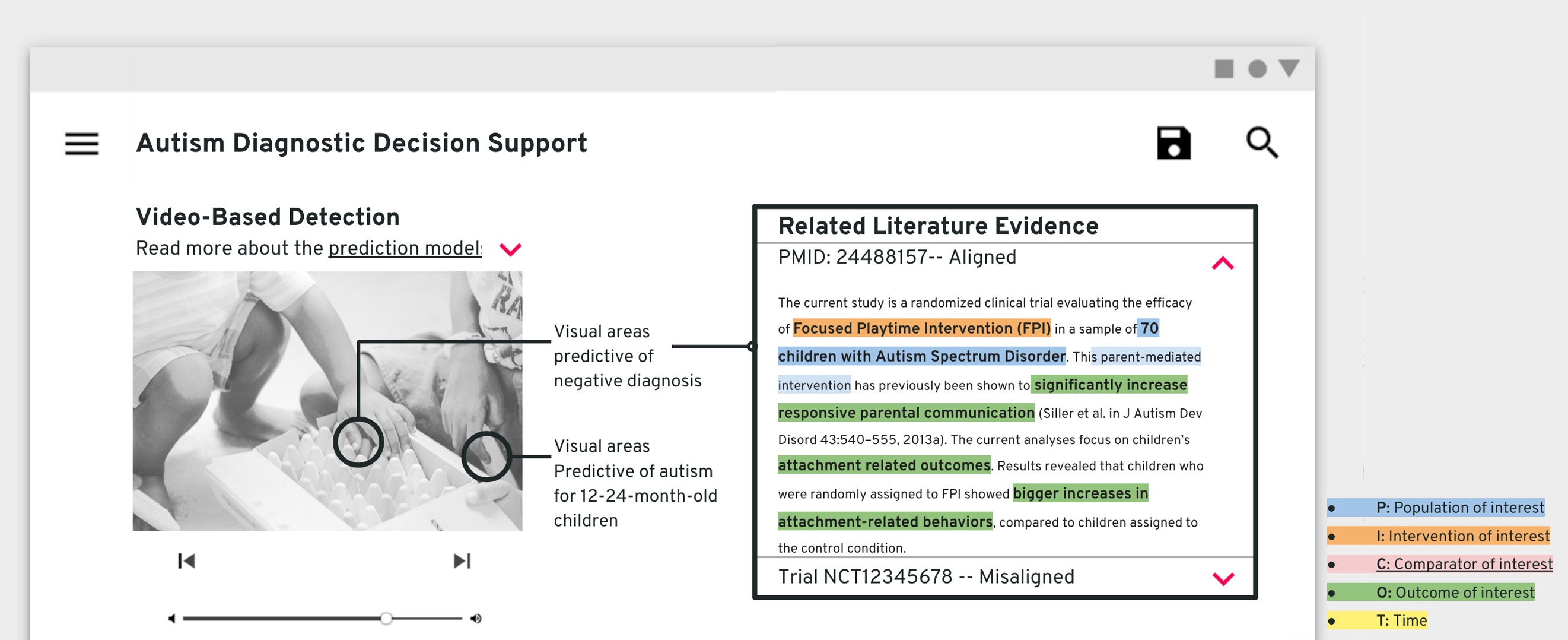}
  \caption{Example interface designs of \sysname~as it aids clinician decision-making alongside other diagnostic or prognostic CDS predictions. The literature contains clinical evidence that can validate or invalidate many CDS predictions, therefore helping clinicians scrutinizing them. This functionality can be particularly valuable for complex systems such as deep-learning-based video analysis because their predictions and predictive mechanism are difficult to explain to clinicians.}
  \label{fig:scenario2}
\end{figure}

Via \sysname~, this work proposed biomedical literature as an alternative approach to providing ``explainability'' to these complex CDS systems, aiding clinicians in scrutinizing the correctness of each prediction. As shown in the example scenario, \textit{to clinicians}, evidence from biomedical literature can be much more easily understandable and intuitively convincing of algorithmic inner-workings. This work opens up a clear design space around supporting otherwise-abstruse CDS predictions with evidence from clinical literature, for example, helping clinicians scrutinizing ML predictions by surfacing the clinical-trial-proven casual relations between ML features and its predicted diagnoses (Figure \ref{fig:scenario2}.)

\paragraphbe{Open research opportunity:} Designing a better blend of heterogeneous clinical decision supports. A key premise of this work is to explore what an AI-\textit{infused} clinical decision-making process might look like in future healthcare. While research has created numerous CDS systems, most focused on one system, one clinical decision, and one clinical domain. However, clinical decision-making is a continuous and iterative process; It consists of a series of micro-decisions. These micro-decisions are often cross-modal and cross-disciplinary, therefore involving distinct CDS systems and potential risks. We encourage future research to investigate how heterogeneous AI decision supports (e.g., literature-based and EHR-based) can best collaborate with clinician teams, forming an effective multi-AI, multi-clinician team. This example scenario offers a small first step towards this ambitious goal.

\section{Closing Notes}
Abstruse learning algorithms and complex datasets increasingly characterize modern decision support system.
As a result, clinicians cannot easily or rapidly scrutinize the CDSS recommendation when facing a difficult diagnosis or treatment decision in practice.
Over-trusting or under-trusting CDSS recommendations are frequent, leading to preventable diagnostic or treatment errors.
Prior research has explored supporting such assessments by explaining DST data inputs and algorithmic mechanisms.
This paper explores a different approach: By providing precisely relevant, scientific evidence from biomedical literature. 
We present a proof-of-concept system, \sysname, to demonstrate the technical and design feasibility of this approach across three domains (cardiovascular diseases, autism, cancer).
It can effectively identify clinical trial reports based on lengthy clinical questions (e.g., "risks of catheter infection among adult patients in intensive care unit who require arterial catheters, if treated with povidone iodine-alcohol").
This capability enables the system to identify clinical trials relevant to diagnostic/treatment hypotheses -- a clinician's or a DST's.
Further, \sysname\ can identify key parts of a clinical trial abstract, including patient population (e.g., adult patients in intensive care unit who require arterial catheters), intervention (povidone iodine-alcohol), and outcome (risks of catheter infection).
Through two example use scenarios of the system, we have demonstrated the many design opportunities and open research questions that this capability opens up.

At a higher level, this work proposes a future where intelligent literature tools can serve as a decision support infrastructure and support many clinical decisions across domains. Such an information infrastructure should be valuable both independently (as illustrated in use scenario 1) and when supporting other intelligent systems, particularly for practitioners in rural or low-resource hospitals where data-intensive CDS is less available (scenario 2). A decision-support infrastructure -- because it operates at a PubMed scale -- can have an outsized impact on clinical practice and improving the quality of patient care.
\bibliographystyle{ACM-Reference-Format}
\bibliography{ref/bioNLP,ref/healthcare,ref/sensemaking,ref/path}

\appendix

\end{document}